\documentclass[sigconf]{acmart}
\AtBeginDocument{%
  }

\copyrightyear{2023}
\acmYear{2023}
\setcopyright{acmlicensed}\acmConference[MM '23]{Proceedings of the 31st
ACM International Conference on Multimedia}{October 29-November 3,
2023}{Ottawa, ON, Canada}
\acmBooktitle{Proceedings of the 31st ACM International Conference on
Multimedia (MM '23), October 29-November 3, 2023, Ottawa, ON, Canada}
\acmPrice{15.00}
\acmDOI{10.1145/3581783.3612011}
\acmISBN{979-8-4007-0108-5/23/10}

\acmSubmissionID{1399}

\usepackage{balance}
\usepackage{amsmath}
\usepackage{amsthm}
\usepackage{algorithm}
\usepackage{algorithmic}
\usepackage{bm}
\usepackage{makecell}
\usepackage{subfigure}

\begin{document}

\title{UER: A Heuristic Bias Addressing Approach \\for Online Continual Learning}


\author{Huiwei Lin}
\affiliation{%
  \institution{Harbin Institute of Technology}
  \city{Shenzhen}
  \country{China}}
\email{linhuiwei@stu.hit.edu.cn}

\author{Shanshan Feng}
\authornote{Corresponding author: Shanshan Feng}
\affiliation{%
  \institution{Harbin Institute of Technology}
  \city{Shenzhen}
  \country{China}}
\email{victor_fengss@foxmail.com}

\author{Baoquan Zhang}
\affiliation{%
  \institution{Harbin Institute of Technology}
  \city{Shenzhen}
  \country{China}}
\email{zhangbaoquan@stu.hit.edu.cn}

\author{Hongliang Qiao}
\affiliation{%
  \institution{Harbin Institute of Technology}
  \city{Shenzhen}
  \country{China}}
\email{21s151112@stu.hit.edu.cn}

\author{Xutao Li}
\affiliation{%
  \institution{Harbin Institute of Technology}
  \city{Shenzhen}
  \country{China}}
\email{lixutao@hit.edu.cn}

\author{Yunming Ye}
\affiliation{%
  \institution{Harbin Institute of Technology}
  \city{Shenzhen}
  \country{China}}
\email{yeyunming@hit.edu.cn}

\renewcommand{\shortauthors}{Huiwei Lin et al.}

\begin{abstract}
  Online continual learning aims to continuously train neural networks from a continuous data stream with a single pass-through data. As the most effective approach, the rehearsal-based methods replay part of previous data. Commonly used predictors in existing methods tend to generate biased dot-product logits that prefer to the classes of current data, which is known as a bias issue and a phenomenon of forgetting. Many approaches have been proposed to overcome the forgetting problem by correcting the bias; however, they still need to be improved in online fashion. In this paper, we try to address the bias issue by a more straightforward and more efficient method. By decomposing the dot-product logits into an angle factor and a norm factor, we empirically find that the bias problem mainly occurs in the angle factor, which can be used to learn novel knowledge as cosine logits. On the contrary, the norm factor abandoned by existing methods helps remember historical knowledge. Based on this observation, we intuitively propose to leverage the norm factor to balance the new and old knowledge for addressing the bias. To this end, we develop a heuristic approach called unbias experience replay (UER). UER learns current samples only by the angle factor and further replays previous samples by both the norm and angle factors. Extensive experiments on three datasets show that UER achieves superior performance over various state-of-the-art methods. The code is in \url{https://github.com/FelixHuiweiLin/UER}.
\end{abstract}

\ccsdesc[500]{Computing methodologies~Cross-validation}
\ccsdesc[500]{Computing methodologies~Image representations}

\keywords{neural networks, online continual learning, image classification}


\maketitle

\section{Introduction}
\label{sec:introduction}

At present, deep neural networks~\cite{nie2022search,nie2019multimodal,nie2017enhancing} still cannot have the conventional ability of continuous learning as human beings. If the model adapts to new data through fine-tuning, it will significantly degrade the performance of old data. Such a phenomenon is named catastrophic forgetting (CF)~\cite{delange2021continual}. Since solving this problem is of great significance, continual learning~\cite{liu2021lifelong,liu2022ai,liu2020lifelong} has been proposed to train the model to achieve knowledge accumulation without forgetting. Particularly, online continual learning (OCL) is a realistic scenario of continual learning. It requires overcoming the CF problem of continuously learning from an infinite and non-stationary data stream, where the data can be accessed only once. 

The rehearsal-based methods have shown superior performance for OCL among all approaches of continual learning~\cite{lin2022anchor}. This family of methods~\cite{yin2021mitigating,mai2022online, lin2023pcr} set an episodic memory buffer to store part of previous samples and replay them in the learning process of current samples. In this type of methods, the phenomenon of CF is exhibited as a biased issue of prediction due to data imbalance. Specifically, the number of current samples in the data stream is generally greater than the number of previous samples in the fixed buffer. Hence, the model tends to score the samples with biased predictions and easily classifies most samples to current classes.

\begin{figure*}[t]
\centering
\includegraphics[scale=0.6]{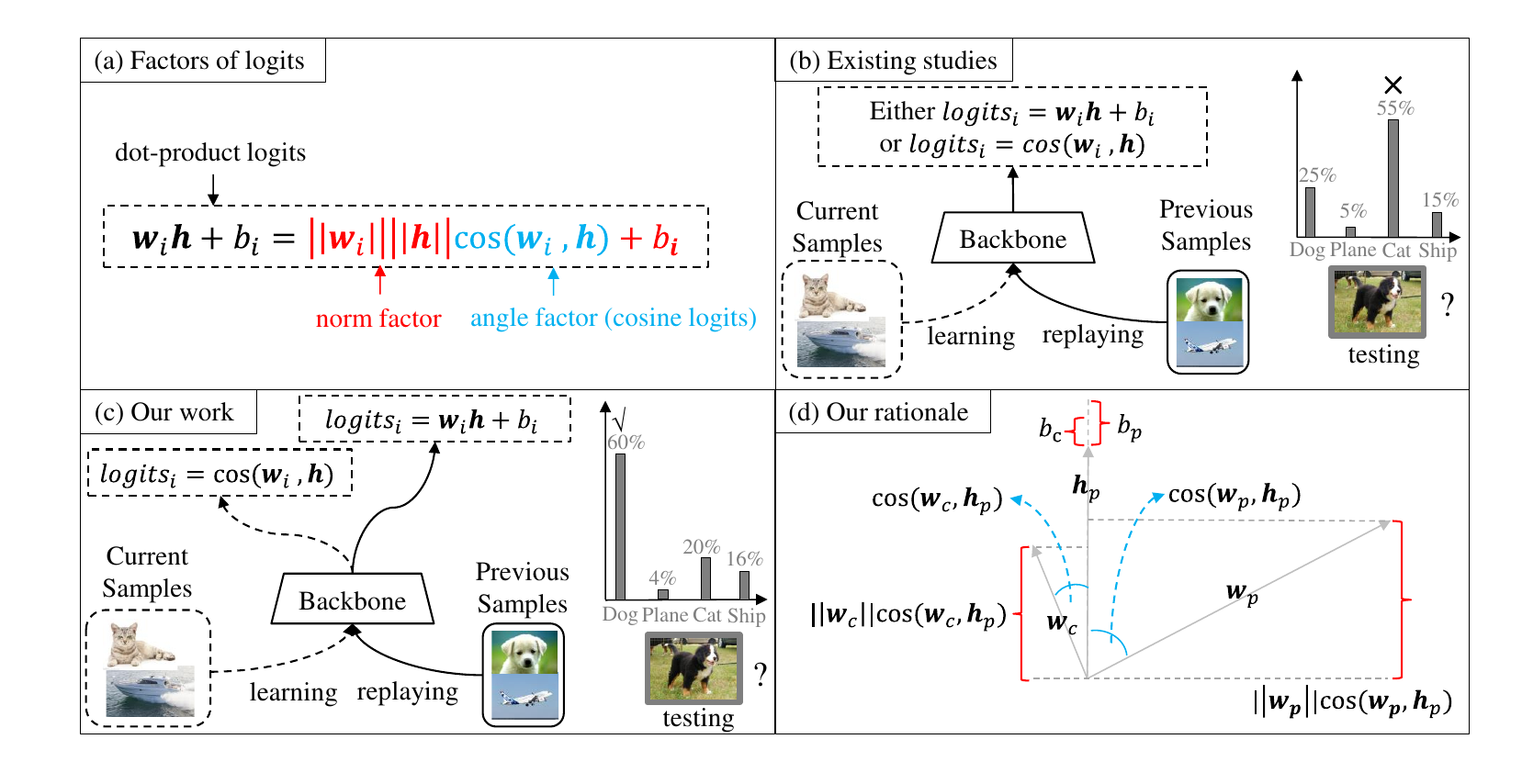}
\vspace{-2ex}
\caption{Illustration of our work. (a) The dot-product logits can be decomposed into a norm factor and an angle factor; (b) Existing studies, which learn all samples by calculating logits in the same way; (c) Different from existing studies, the model in our work learns current samples only by the angle factor meanwhile replays previous samples by both the angle factor and the norm factor. (d) By leveraging the norm factor, the model can correctly classify samples.}
\label{fig:problem}%
\vspace{-3ex}
\end{figure*}

Although existing methods mitigate the forgetting problem by correcting the bias, they still need to be improved in online fashion. Specifically, there are three main challenges that need to be addressed: 1) \textbf{Real-time ability}. Some methods~\cite{wu2019large,zhao2020maintaining,mai2021supervised} require additional training steps or post-operations, which is not beneficial to the real-time prediction of the model. 2) \textbf{Flexibility}. Some methods~\cite{ahn2021ss,caccia2021new} rely heavily on the clear boundary between the old and new classes, and they cannot be extended from the predictor to the feature
extractor. 3) \textbf{Adjustability}. Other methods~\cite{chrysakisonline,caccia2021new} are difficult to adjust the scale of bias correction, easily limiting the generalization ability of the models. Thus, a meaningful question arises: Is there a more straightforward and efficient approach that can simultaneously meet these three challenges and effectively address the bias issue for OCL?

To answer this question, we dissect existing rehearsal-based methods and empirically find that the biased predictions (dot-product logits) mainly changes in two parts, as shown in Figure~\ref{fig:problem}(a). Inspired by~\cite{liu2018decoupled}, we decompose the dot-product logits ($\bm{w}\bm{h}+b$) into a norm factor ($\Vert\bm{w}\Vert\Vert\bm{h}\Vert+b$) and an angle factor ($cos(\bm{w},\bm{h})$). The unbalance data leads to an unbalanced competition of gradient propagation, which is a major cause of the CF phenomenon, and this effect influences the logits in both factors. Further experiments reveal that these two factors act differently during the training process. On one hand, the angle factor suffers from a serious bias issue in prediction. It can be independently utilized as cosine logits that mainly captures novel knowledge from current samples. On the other hand, the norm factor, which is abandoned by existing works~\cite{hou2019learning,caccia2021new}, is helpful in keeping historical knowledge of previous samples.

Based on this observation, it is feasible to leverage the norm factor to balance old and new knowledge and address the bias issue. Comparing with the angle factor, the norm factor can effectively store historical knowledge, which is helpful to deal with the bias problem. In other words, the dot-product logits, composed of two factors, is suitable for replaying previous samples. Conversely, the cosine logits, based on the angle factor, is made available for learning current samples since it can quickly learn novel knowledge of current samples. Different from existing studies (Figure~\ref{fig:problem}(b)) that calculate logits in the same way, we try to (Figure~\ref{fig:problem}(c)) learn current samples by cosine logits and further replay previous samples by dot-product logits. As shown in Figure~\ref{fig:problem}(d), a sample $\bm{h}_p$ will be mistakenly classified to current class $c$ by cosine logits that satisfies $cos(\bm{w}_c,\bm{h}_p)>cos(\bm{w}_p,\bm{h}_p)$. However, with the help of the norm factor, the model can correctly classify $\bm{h}_p$ to previous class $p$ by dot-product logits that satisfies $\Vert\bm{h}_p\Vert\Vert\bm{w}_c\Vert cos(\bm{w}_c,\bm{h}_p)+b_c<\Vert\bm{h}_p\Vert\Vert\bm{w}_p\Vert cos(\bm{w}_p,\bm{h}_p)+b_p$. Empirical experiments demonstrate that this approach effectively solves the bias problem and achieves better performance.

With these inspirations, we propose unbias experience replay (UER), a heuristic bias addressing approach alleviating the phenomenon of CF for OCL. Its process can be mainly divided into three
components.: 1) The \textbf{learning component} is the basic component that trains current samples from the data stream by cosine logits. It aims to save novel knowledge in the angle factor. 2) To solidify historical knowledge, the \textbf{replaying component} mainly trains previous samples in the memory buffer using both the norm and angle factors, leveraging the norm factor to correct the biased prediction. 3) With the help of the norm factor, the \textbf{testing component} tends to predict unbiased categorical distributions for all samples. Such a simple yet effective method can be used for real-time prediction (\textbf{Real-time ability}), extended to the feature extractor (\textbf{Flexibility}), and the scale of bias addressing is easily adjusted (\textbf{Adjustability}).

Our main contributions can be summarized as follows:
\begin{itemize}
\item[1)] We theoretically analyze the characteristics of dot-product logits and cosine logits, discovering their coupling is beneficial. To the best of our knowledge, this work is the first to combine these two scoring ways for the OCL.

\item[2)] We develop a novel rehearsal-based framework called UER to mitigate the forgetting problem by addressing the bias issue of prediction. The core is the training mechanism that learns novel knowledge of current samples by the angle factor and further enhances historical knowledge of previous samples by the norm factor. 

\item[3)] We conduct extensive experiments on three real-world datasets, and the results consistently demonstrate the superiority of UER over various state-of-the-art methods.
\end{itemize}

\section{Related Work}
\label{sec:relatedwork}

\subsection{Continual Learning}

Recent advances on continual learning can be grouped into three categories. 1) Architecture-based methods divide each stage into a set of specific parameters of the model, containing dynamical network~\cite{rusu2016progressive} and static network~\cite{miao2021continual}. 2) Regularization-based methods take historical knowledge as the prior information of learning novel knowledge, extending the loss function with additional regularization term~\cite{kirkpatrick2017overcoming,dhar2019learning}. 3) Rehearsal-based methods set a fixed-size memory buffer~\cite{lopez2017gradient,chaudhry2018efficient,farajtabar2020orthogonal,tang2021layerwise} or generative model~\cite{cui2021deepcollaboration} to store, produce and replay previous samples with current ones. The kind of method~\cite{cha2021co2l,chaudhry2021using,liu2021rmm,wang2021memory} that saves previous samples in the buffer is still the most effective for knowledge anti-forgetting at present~\cite{buzzega2021rethinking}. Different from them, our method aims at overcoming the forgetting problem for OCL, which requires high real-time performance. It is more difficult than general continual learning setting.

\subsection{Online Continual Learning}
Rehearsal-based methods based on Experience replay (ER)~\cite{rolnick2019experience} are the core solution of OCL. Some approaches~\cite{aljundi2019online,shim2021online} exploit memory retrieval strategy to select more valuable samples from memory. Meanwhile, some approaches~\cite{aljundi2019gradient,jin2021gradient,he2021online} focus on saving more effective samples, belonging to the family of memory update strategy. The others~\cite{mai2021supervised,yin2021mitigating,caccia2021new,guo2022online,gu2022not} called model update strategy improve the training process to learn efficiently. The proposed UER is a novel model update strategy to alleviate the phenomenon of CF by addressing the bias of prediction. Different from existing strategies, UER calculates logits in different ways for current samples and previous samples. Such a training mechanism makes use of the properties of the norm and angle factors. It can be simultaneously used for real-time prediction without additional post-operations, extended from the predictor to the feature extractor, and easy to control the scale of bias addressing.

\section{Rethinking Predictor}
\label{sec:preliminary}

\subsection{Problem Definition}
OCL divides a data stream into a sequence of mini learning batches as $\mathcal{D}=\{\mathcal{B}_t\}_{t=1}^T$, where $\mathcal{B}_t=\{\mathcal{X}_t\times \mathcal{Y}_t\}$ contains the samples $\mathcal{X}_t$ and corresponding labels $\mathcal{Y}_t$. All of learned classes are denoted as $\mathcal{C}_{t}$ until step $t$. The neural network is made up of a feature extractor $\bm{h}=h_{\bm{\Phi}}(\bm{x})$ and a predictor $f(\bm{h})=\bm{W}\bm{h}^T+\bm{b}$, where $\bm{h}$ is non-negative by Relu, $\bm{W}=[\bm{w}_i]$, and $\bm{b}=[b_i]$. Since the output of predictor is dot-product logits, the categorical prediction of sample $\bm{x}$ is $p^{dot}(\bm{x})=[p_i^{dot}]$, and $p_i^{dot}$ is the probability that a sample belongs to class $i\in\mathcal{C}_{t}$:
\begin{equation}
    \label{eq:probability}
	p_i^{dot}=\frac{exp({\bm{w}_ih_{\bm{\Phi}}(\bm{x})^T+b_i})}{\sum_{j\in\mathcal{C}_{t}}exp({\bm{w}_jh_{\bm{\Phi}}(\bm{x})^T+b_j})}.
\end{equation}
In the training process, the model can only access ${B_t}$ and each sample can be seen only once. Its objective function is calculated as
\begin{equation}
    \label{eq:oclloss}
	L_{ocl}=E_{(\bm{x},y)\sim{\mathcal{B}_t}}[l(p^{dot}(\bm{x}), \bm{y})],
\end{equation}
where $l(\cdot)$ is the cross-entropy loss function.

ER~\cite{rolnick2019experience}, as a classic rehearal-based method, allocates a memory buffer $\mathcal{M}$ to store part of previous samples, and selects them as $\mathcal{B}_{\mathcal{M}}$ to replay with current samples. Thus, the objective function can be changed to
\begin{equation}
    \label{eq:erloss}
	L_{er}=E_{(\bm{x},y)\sim{\mathcal{B}_t\cup\mathcal{B}_{\mathcal{M}}}}[l(p^{dot}(\bm{x}), \bm{y})].
\end{equation}

LUCIR~\cite{hou2019learning} is an improved method of ER. Generally, the dot-product logits ($\bm{w}\bm{h}+b$) of predictor can be parameterized as the norm factor ($\Vert\bm{w}\Vert\Vert\bm{h}\Vert+b$) and the angle factor ($cos(\bm{w},\bm{h})$)~\cite{liu2018decoupled}. In LUCIR, the norm factor is considered as an important part with bias. Hence, the angle factor is used separately as cosine logits to calculate the categorical probability prediction $p^{cos}(\bm{x})=[p_i^{cos}]$. And its objective function is 
\begin{equation}
    \label{eq:lucirloss}
    L_{lucir}=E_{(\bm{x},y)\sim{\mathcal{B}_t\cup\mathcal{B}_{\mathcal{M}}}}[l(p^{cos}(\bm{x}), \bm{y})].
\end{equation}
where 
\begin{equation}
    \label{eq:probability_cos}    p_i^{cos}=\frac{exp({cos(\bm{w}_i,h_{\bm{\Phi}}(\bm{x}))\cdot\gamma})}{\sum_{j\in\mathcal{C}_{1:t}}exp({cos(\bm{w}_j,h_{\bm{\Phi}}(\bm{x}))\cdot\gamma})}.
\end{equation}
$\gamma$ is a scale parameter and usually set as 10.

\begin{table*}[t]
\renewcommand\tabcolsep{2pt}
\centering
\caption{The analysis results of different ways to train and test on Split CIFAR100 (Buffer Size=5000). It reports the final accuracy of all classes ($A_{all}$), previous classes ($A_{p}$), and current classes ($A_{c}$). Meanwhile, it also reports the average norm value of previous ($\Vert\bm{W}_p\Vert$) and current classes ($\Vert\bm{W}_c\Vert$), the average weight value of previous (Mean($\bm{W}_p$)) and current classes (Mean($\bm{W}_p$)), and the average bias value of previous (Mean($b_p$)) and current classes (Mean($b_c$)).}
\vspace{-2ex}
\label{table:analysis}
\begin{tabular}{lccc|ccc|cccc|cc}
\hline
\multicolumn{1}{l|}{Index}                     & Learning                      & Replaying                     & Testing                    & $A_{all}$  & $A_p$   & $A_c$                 & $\Vert\bm{W}_p\Vert$ & $\Vert\bm{W}_c\Vert$    & Mean($\bm{W}_p$) & Mean($\bm{W}_c$) & Mean($b_p$) & Mean($b_c$)\\ \hline
\multicolumn{1}{l|}{1 (ER)}    & dot-product                          & dot-product                          & dot-product                          & 27.3 & 17.7 & 36.9 & 2.08  & 1.86  & 0.0044± 0.0274 & -0.0038± 0.0221 & -0.22 & 0.22\\
\multicolumn{1}{l|}{2}   & dot-product                          & dot-product                          & cosine & 24.1  & 12.7  & 35.6 & 2.08  & 1.86  & 0.0044± 0.0274 & -0.0038± 0.0221 & -0.22 & 0.22\\
\multicolumn{1}{l|}{3 (LUCIR)} & cosine & cosine & cosine & 26.9 & 14.9 & 38.9 & 2.00  & 1.64  & 0.0398±0.0235 & 0.0471± 0.0146 & -0.00 & 0.00\\
\multicolumn{1}{l|}{4}   & cosine & cosine & dot-product                          & 28.2 & 28.8 & 27.6 & 2.00  & 1.64  & 0.0398±0.0235 & 0.0471± 0.0146 & -0.00 & 
0.00\\
\multicolumn{1}{l|}{5}   & dot-product & cosine                          & dot-product                          & 19.6 & 0.0 & 39.1 & 1.37  & 2.09  & 0.0197±0.0115 & 0.0219± 0.0270 & -1.03 & 1.03\\
\multicolumn{1}{l|}{6}   & dot-product & cosine                          & cosine                          & 27.3 & 25.3 & 29.4 & 1.37  & 2.09  & 0.0197±0.0115 & 0.0219± 0.0270 & -1.03 & 1.03\\
\multicolumn{1}{l|}{7}   & cosine & dot-product                          & cosine                          & 21.3 & 0.5 & 42.2 & 2.52  & 1.51  & 0.0508±0.0371 & 0.0565± 0.0112 & 0.20 & -0.20\\
\multicolumn{1}{l|}{\textbf{8 (Ours)}}   & \textbf{cosine} & \textbf{dot-product}                          & \textbf{dot-product}                          & \textbf{29.5} & \textbf{30.9} & \textbf{28.1} & \textbf{2.52}  & \textbf{1.51}  & \textbf{0.0508±0.0371} 
 & \textbf{0.0565± 0.0112}& \textbf{0.20} & \textbf{-0.20}\\\hline
\end{tabular}
\vspace{-2ex}
\end{table*}

\begin{figure*}[t]
\centering
\subfigure[The accumulate changes of parameters during the learning process.]{
    \begin{minipage}[t]{0.3\linewidth}
        \centering
        \includegraphics[scale=0.14]{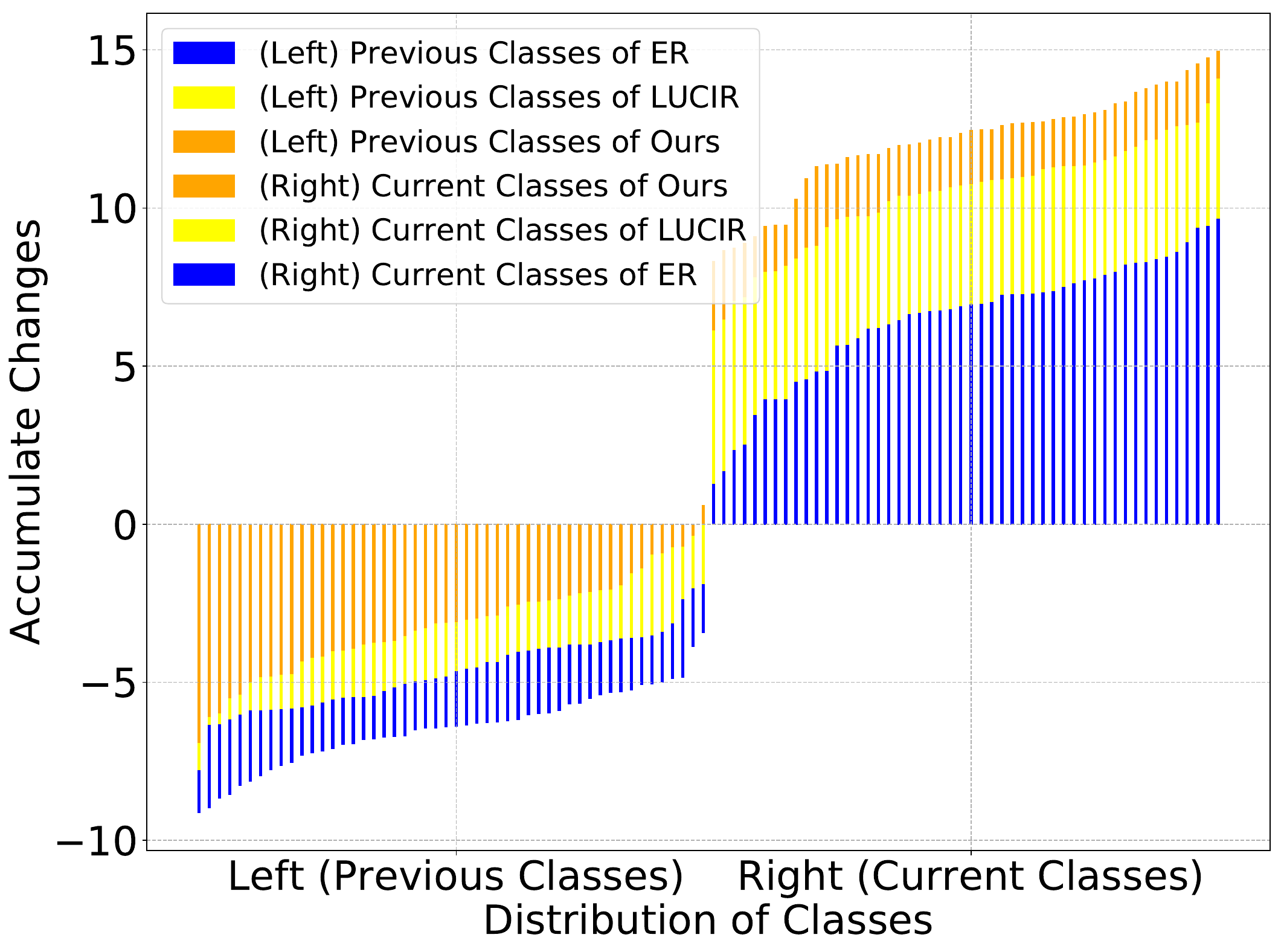}
    \end{minipage}
}
\hspace{.10in}
\subfigure[The average norm value of all classes.]{
    \begin{minipage}[t]{0.3\linewidth}
        \centering
        \includegraphics[scale=0.14]{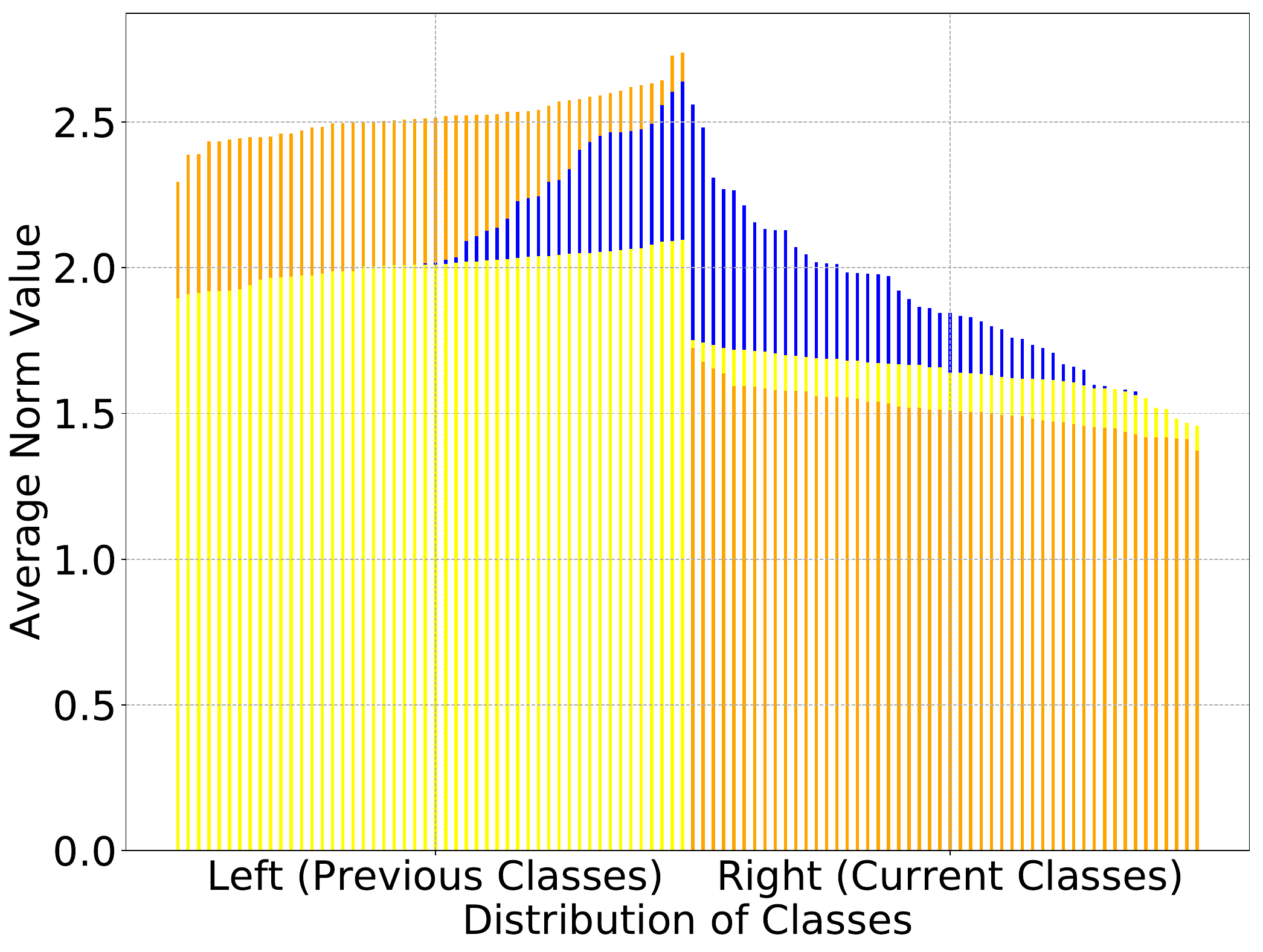}
    \end{minipage}
}
\hspace{.10in}
\subfigure[The average categorical probability of testing samples.]{
    \begin{minipage}[t]{0.3\linewidth}
        \centering
        \includegraphics[scale=0.14]{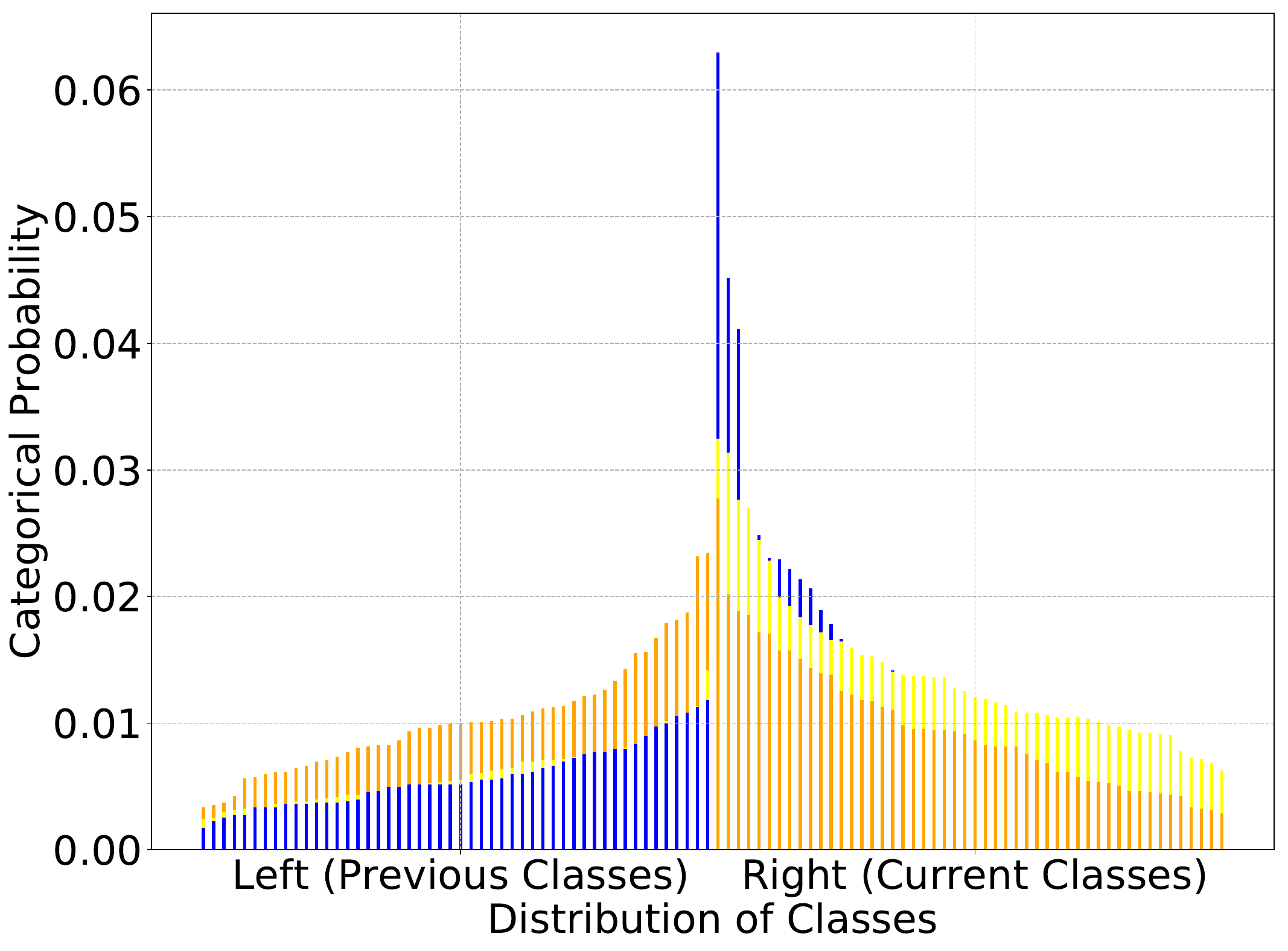}
    \end{minipage}
}
\centering
\vspace{-4ex}
\caption{The analysis results of different ways on Split CIFAR100 when the memory buffer size is 5000. }
\label{fig:analysis}
\vspace{-2ex}
\end{figure*}

\subsection{Analysis of Catastrophic Forgetting}
There is a competition between samples of each class in the process of gradient propagation. By chain rule, the gradient of predictor for a training sample $\bm{x}$ of class $y$ can be expressed as
\begin{equation}
    \label{eq:gradient_w}
    \frac{\partial L_{er}}{\partial \bm{w}_j}=\left\{
    \begin{aligned}
        (p_y^{dot}-1)\bm{h}&,&j= y \\
        p_j^{dot}\bm{h}&,&j\neq y
    \end{aligned}
    \right.,
\end{equation}
and
\begin{equation}
    \label{eq:gradient_b}
    \frac{\partial L_{er}}{\partial b_j}=\left\{
    \begin{aligned}
        p_y^{dot}-1&,&j= y \\
        p_j^{dot}&,&j\neq y
    \end{aligned}
    \right..
\end{equation}
With a positive learning rate $\eta$ and non-negative feature $\bm{h}$, it provides the positive change for associated parameters of class $y$ as $\bm{w}_y=\bm{w}_y-\eta(p_y-1)\bm{h}$ and $b_y=b_y-\eta (p_y-1)$, and propagates the negative changes to the other parameters as $\bm{w}_j=\bm{w}_j-\eta p_j\bm{h}$ and $b_j=b_j-\eta p_j$. It means that not only the values of $\bm{w}_y$ and $b_y$ will become larger, but also the angles of $\bm{w}_y$ and $\bm{h}$ will be smaller. Thus, it increases the logits score of the $y$-th dimension as well as decreases the logits scores of other dimensions. And the process is the same when using cosine logits by
\begin{equation}
    \label{eq:gradient_w_cos}
    \frac{\partial L_{lucir}}{\partial \bm{w}_j}=\left\{
    \begin{aligned}
        (p_y^{cos}-1)\hat{\bm{h}}&,&j= y \\
        p_j^{cos}\hat{\bm{h}}&,&j\neq y
    \end{aligned}
    \right.,
\end{equation}
where
\begin{equation}
    \label{eq:h}
	\hat{\bm{h}}=\frac{\frac{\bm{h}}{\Vert\bm{h}\Vert}-cos(\bm{w},\bm{h})\frac{\bm{w}}{\Vert\bm{w}\Vert}}{\Vert\bm{w}\Vert}.
\end{equation}

The biased issue is a phenomenon of CF, mainly caused by the unbalanced competition of gradient propagation. Generally, the number of samples belonging to previous classes in the fixed buffer will decrease as the learning process continues. And the number of samples belonging to current classes remains unchanged. If the model incrementally learns by Eq.~\ref{eq:erloss} or Eq.~\ref{eq:lucirloss}, the associated parameters of current classes receive more positive changes while the others obtain more negative changes. It means the gradients of previous classes lose their competitiveness. Besides, the logits of current classes tend to be larger than the previous ones. Hence, the model is more likely to classify most samples to the classes of current samples, reducing the performance of previous classes.

\subsection{Analysis of Biased Predictor}
The unbalanced gradient mainly affects the predictor from the angle and norm factors. The norm factor accounts for intra-class variation, and the angle factor accounts for inter-class variation~\cite{liu2018decoupled}. To study their changes in OCL, we conduct experiments on CIFAR100 with two incremental stages and 50 classes per stage. We set two general ways to calculate logits in the experiment: dot-product logits (both factors) and cosine logits (only angle factor). In Table~\ref{table:analysis}, we take different ways to calculate logits, and record their performances.

Experiment results show that the characteristics of these two factors differ from each other. On one hand, the bias issue mainly occurs in the angle factor, which efficiently learns novel knowledge. Comparing index 1 and 2, replacing dot-product logits with cosine logits in the testing phase leads to worse final accuracy, particularly for previous classes. Meanwhile, the method in index 3 performs better for current classes but worse for previous classes compared to the one in index 1. On the other hand, the norm factor, which has been abandoned in existing studies~\cite{hou2019learning}, benefits previous classes. As seen in index 3 and 4, the final performance is improved if cosine logits is changed to dot-product logits in the testing phase.

With this observation, the two factors can be utilized in different occasions. The angle factor is effective in adapting novel knowledge with inter-class variation. Consequently, cosine logits that depends on the angle factor is suitable for learning current samples. And dot-product logits can be used for replaying previous samples since it can leverage the norm factor to enhance historical knowledge with intra-class variation. Motivated by these findings, we propose a method that learns current samples by cosine logits, replays previous samples, and predicts testing samples by dot-product logits.

\subsection{Analysis of Proposed Method}
In addition to the results in Table~\ref{table:analysis}, we present more detailed results for ER (blue), LUCIR (yellow) and our method (orange) in Figure~\ref{fig:analysis}. We calculate the difference between the parameters of predictor before and after learning current classes, taking it as the accumulated changes of parameter as shown in Figure~\ref{fig:analysis} (a). Meanwhile, we report the average norm value of weight vector for all classes in Figure~\ref{fig:analysis} (b). Moreover, we calculate the average categorical probability of all testing samples by the latest model, illustrated in Figure~\ref{fig:analysis} (c). Further analysis and experiment results verify the feasibility of our proposed method.

Firstly, learning current samples using cosine logits can save the novel knowledge into the angle factor. In index 7 and 8, the performance on previous classes is almost zero when testing with cosine logits. The reason behind this is that the angular similarity is almost the determining factor for gradients in Equation (\ref{eq:gradient_w_cos}). For one thing, all vectors in Equation (\ref{eq:gradient_w_cos}) are normalized and it greatly enhances the influence of the angular similarity. For another thing, the change of $\bm{w}$ is constrained by $\hat{\bm{h}}$, as large $\Vert\bm{w}\Vert$ can decrease gradients so much that the weights pass is not able to update effectively. Moreover, the dot-product logits is associated with a value range $[-\infty,+\infty]$, while the value range of cosine logits is $[-1,1]$. It causes the cosine logits to produce smaller change for previous classes and larger change for current classes (as seen in the blue and orange parts in Figure~\ref{fig:analysis} (a)). And it also raises the norm factor of previous classes and reduces the norm factor of current classes (as observed in the blue and orange parts in Figure~\ref{fig:analysis} (b)), which is helpful in learning novel knowledge by the angle factor.

Secondly, replaying previous samples with dot-product logits can enhance historical knowledge through the norm factor. Comparing index 7 and 8, using dot-product logits in the testing process results in the best performance for previous classes, even surpassing the performance of current classes. Different from cosine logits, the gradient in Equation (\ref{eq:gradient_w}) depends on both two factors. In the meantime, the gradient of Equation (\ref{eq:gradient_w}) is only associated with previous samples in the buffer. Since the angle factor is easily influenced by unbalanced gradient, this process tends to pay more attention to the norm factor. As a result, replaying by dot-product logits not only decreases the changes for previous classes but also increases the changes for currents classes (as seen in the yellow and orange parts in Figure~\ref{fig:analysis} (a)). In the meantime, it raises the norm factor of previous classes as well as reduces the norm factor of current classes (indicated by the yellow and orange parts in Figure~\ref{fig:analysis} (a)). It causes  that $\Vert\bm{W}_p\Vert$ is larger than $\Vert\bm{W}_c\Vert$, and Mean($b_c$) is smaller than Mean($b_p$). Although the angle factor suffers from $cos(\bm{w}_c,\bm{h})>cos(\bm{w}_p,\bm{h})$, it can be leveraged by $\Vert\bm{w}_p\Vert\Vert\bm{h}\Vert+b_p>\Vert\bm{w}_c\Vert\Vert\bm{h}\Vert+b_c$, as we described in Figure~\ref{fig:problem} (d).

As shown in Table~\ref{table:analysis}, our proposed method achieves the best performance among all combinations of different logits. Based on the combination of different logits, our method quickly learns novel knowledge and effectively keep historical knowledge. As shown in Figure~\ref{fig:analysis} (c), the posterior probability distribution of our method is relatively balanced, whereas those of ER and LUCIR tend to favor the current classes. Therefore, our motivation of learning current samples
by cosine logits while replaying previous samples by dot-product logits is reasonable and practicable.

\section{Unbias Experience Replay}
\label{sec:method}
Inspired by these analyses, we propose a novel UER framework, which is a simple but effective solution for OCL. The framework consists of a CNN-based feature extractor and a predictor. The key idea is to mitigate the forgetting problem of the model by addressing the bias issue of probability distribution. The model can quickly learn novel knowledge of current samples by the angle factor, and effectively consolidate the historical knowledge learned from previous samples by both two factors. The overall workflow can be divided into three main modules: memory buffer operation, online continual training and online continual testing. 

\subsection{Memory Buffer Operation}
Conveniently, the memory buffer in our framework has a fixed size, no matter how large the amount of samples is. On the one hand, it updates the memory buffer by reservoir sampling strategy. When the memory cannot load all samples, the reservoir sampling algorithm can ensure that the probability of each sample being extracted is equal. On the other hand, it randomly retrieves previous samples from the memory buffer to participate in the learning of current samples.

\begin{table*}[t]
\renewcommand\tabcolsep{3pt}
\centering
\caption{Final Average Accuracy Rate (higher is better). The best scores are in boldface, and the second best scores are underlined.}
\vspace{-2ex}
\label{tableaccuracy}
\begin{tabular}{l|cccc|cccc|cccc}
\hline
Methods & \multicolumn{4}{c|}{Split CIFAR10} & \multicolumn{4}{c|}{Split CIFAR100}& \multicolumn{4}{c}{Split MiniImageNet}\\ \hline
Buffer & \multicolumn{1}{c|}{100} & \multicolumn{1}{c|}{200} & \multicolumn{1}{c|}{500} & \multicolumn{1}{c|}{1000}    & \multicolumn{1}{c|}{500} & \multicolumn{1}{c|}{1000} & \multicolumn{1}{c|}{2000} & \multicolumn{1}{c|}{5000}     &\multicolumn{1}{c|}{500} & \multicolumn{1}{c|}{1000} & \multicolumn{1}{c|}{2000} & 5000     \\ \hline
IID          & \multicolumn{4}{c|}{58.1\scriptsize±2.5}                                  & \multicolumn{4}{c|}{17.3\scriptsize±0.8}                                     & \multicolumn{4}{c}{18.2\scriptsize±1.1}                                     \\
IID++~\cite{caccia2021new}          & \multicolumn{4}{c|}{64.2\scriptsize±2.1}                                  & \multicolumn{4}{c|}{23.5\scriptsize±0.8}                                     & \multicolumn{4}{c}{20.7\scriptsize±1.0}                                     \\
FINE-TUNE          & \multicolumn{4}{c|}{17.9\scriptsize±0.4}                                  & \multicolumn{4}{c|}{5.9\scriptsize±0.2}                                     & \multicolumn{4}{c}{4.3\scriptsize±0.2}                                                                         \\\hline
ER~\cite{rolnick2019experience} (NeurIPS2019) & 33.8\scriptsize±3.2& 41.7\scriptsize±2.8                 & 46.0\scriptsize±3.5                 & 46.1\scriptsize±4.3 &14.5\scriptsize±0.8& 17.6\scriptsize±0.9                  & 19.7\scriptsize±1.6                  & 20.9\scriptsize±1.2 & 11.2\scriptsize±0.6 & 13.4\scriptsize±0.9                   & 16.5\scriptsize±0.9                  & 16.2\scriptsize±1.7\\
GSS~\cite{aljundi2019gradient} (NeurIPS2019)& 23.1\scriptsize±3.9& 28.3\scriptsize±4.6                 & 36.3\scriptsize±4.1                 & 44.8\scriptsize±3.6 & 14.6\scriptsize±1.3& 16.9\scriptsize±1.4                  & 19.0\scriptsize±1.8                  & 20.1\scriptsize±1.1 & 10.3\scriptsize±1.5 & 13.9\scriptsize±1.0                   & 14.6\scriptsize±1.1                  & 15.5\scriptsize±0.9\\ \hline
MIR~\cite{aljundi2019online} (NeurIPS2019)& 34.8\scriptsize±3.3& 40.3\scriptsize±3.3                 & 42.6\scriptsize±1.7                 & 47.4\scriptsize±4.1 & 14.8\scriptsize±0.7& 18.1\scriptsize±0.7                  & 20.3\scriptsize±1.6                  & 21.6\scriptsize±1.7 & 11.9\scriptsize±0.6 & 14.8\scriptsize±1.1                   & 17.2\scriptsize±0.8                  & 17.2\scriptsize±1.2\\
ASER~\cite{shim2021online} (AAAI2021) & 33.7\scriptsize±3.7& 31.6\scriptsize±3.4                 & 42.1\scriptsize±3.0                 & 42.3\scriptsize±2.9 & 13.0\scriptsize±0.9& 16.1\scriptsize±1.1                  & 17.7\scriptsize±0.7                  & 18.9\scriptsize±1.0 & 10.5\scriptsize±1.1 & 13.8\scriptsize±0.9                   & 16.1\scriptsize±0.9                  & 18.1\scriptsize±1.1\\ 
GMED~\cite{jin2021gradient} (NeurIPS2021) & 32.8\scriptsize±4.7& 43.6\scriptsize±5.1                 & 52.5\scriptsize±3.9                 & 51.3\scriptsize±3.6 & 15.0\scriptsize±0.9& 18.8\scriptsize±0.7                  & 21.1\scriptsize±1.2                  & 23.0\scriptsize±1.5 & 11.9\scriptsize±1.2 & 15.3\scriptsize±1.3                   & 18.0\scriptsize±0.8                  & 19.6\scriptsize±1.0\\ \hline
A-GEM~\cite{chaudhry2018efficient} (ICLR2019)& 17.5\scriptsize±1.7& 17.4\scriptsize±2.1                 & 17.9\scriptsize±0.7                 & 18.2\scriptsize±1.5 & 5.4\scriptsize±0.6& 5.6\scriptsize±0.5                  & 5.4\scriptsize±0.7                  & 4.6\scriptsize±1.0 & 5.0\scriptsize±1.0 & 4.7\scriptsize±1.1                   & 5.0\scriptsize±2.3                  & 4.8\scriptsize±0.8\\
LUCIR~\cite{hou2019learning} (CVPR2019)   & 40.3\scriptsize±2.0& 46.0\scriptsize±1.4                & 50.6\scriptsize±2.2                 & 55.7\scriptsize±3.9 & 14.5\scriptsize±0.4& 17.4\scriptsize±0.8                &20.1\scriptsize±0.7                  & 21.8\scriptsize±1.3 &  12.4\scriptsize±0.6 & 14.7\scriptsize±0.9                   & 16.9\scriptsize±1.1                  & 18.4\scriptsize±1.3\\
ER-WA~\cite{zhao2020maintaining} (CVPR2020)& 36.9\scriptsize±2.9& 42.5\scriptsize±3.4                 & 48.6\scriptsize±2.7                 & 45.9\scriptsize±5.3 & 18.3\scriptsize±0.7& 21.7\scriptsize±1.2                  & 23.6\scriptsize±0.9                  & 24.0\scriptsize±1.8 & 15.1\scriptsize±0.7 & 17.1\scriptsize±0.9                   & 18.9\scriptsize±1.4                  & 18.5\scriptsize±1.5\\
DER++~\cite{buzzega2020dark} (NeurIPS2020)& 40.9\scriptsize±1.4& 45.3\scriptsize±1.7                 & 52.8\scriptsize±2.2                 & 53.9\scriptsize±1.9 & 15.5\scriptsize±1.0& 17.2\scriptsize±1.1                  & 19.5\scriptsize±1.2                  & 20.2\scriptsize±1.3 & 11.9\scriptsize±1.0 & 14.8\scriptsize±0.7                   & 16.1\scriptsize±1.3                  & 15.5\scriptsize±1.3\\
SS-IL~\cite{ahn2021ss} (ICCV2021) & 36.8\scriptsize±2.1 & 42.2\scriptsize±1.4                 &  44.8\scriptsize±1.6                 & 47.4\scriptsize±1.5 & 19.5\scriptsize±0.6  & 21.9\scriptsize±1.1                  & 24.5\scriptsize±1.4                  & 24.7\scriptsize±1.0 & 18.0\scriptsize±0.7 & 19.7\scriptsize±0.9                   & 21.7\scriptsize±1.0                  & 24.4\scriptsize±1.6 \\
SCR~\cite{mai2021supervised} (CVPR2021) & 35.0\scriptsize±2.9& 45.4\scriptsize±1.0                 & \underline{55.7\scriptsize±1.6}                 & \underline{59.8\scriptsize±1.6} & 13.3\scriptsize±0.6& 16.2\scriptsize±1.3                  & 18.2\scriptsize±0.8                  & 19.3\scriptsize±1.0 & 12.1\scriptsize±0.7 & 14.7\scriptsize±1.9                   & 16.8\scriptsize±0.6                  & 18.6\scriptsize±0.5\\
ER-ACE~\cite{caccia2021new} (ICLR2022)   & \textbf{44.3\scriptsize±1.5} & \textbf{49.7\scriptsize±2.4}                & 54.9\scriptsize±1.4                 & 57.5\scriptsize±1.9& \underline{19.7\scriptsize±0.8} & \underline{23.1\scriptsize±0.8}                &\underline{24.8\scriptsize±0.9}                  & \underline{27.0\scriptsize±1.2} &  \underline{18.1\scriptsize±0.5}& \underline{20.3\scriptsize±1.3}                   & \underline{24.8\scriptsize±1.1}                  & \underline{26.2\scriptsize±1.0} \\
ER-DVC~\cite{gu2022not} (CVPR2022) & 36.3\scriptsize±2.6 & 45.4\scriptsize±1.4                 &  50.6\scriptsize±2.9                & 52.1\scriptsize±2.5 & 16.8\scriptsize±0.8 & 19.7\scriptsize±0.7                  & 22.1\scriptsize±0.9                  & 24.1\scriptsize±0.8 & 13.9\scriptsize±0.6 & 15.4\scriptsize±0.7                   & 17.2\scriptsize±0.8                  & 19.1\scriptsize±0.9 \\
OBC~\cite{chrysakisonline} (ICLR2023) & 40.5\scriptsize±2.1  & 46.4\scriptsize±1.6  & 53.4\scriptsize±2.3  & 55.3\scriptsize±2.7 & 18.7\scriptsize±0.9  & 22.1\scriptsize±0.6  & 24.0\scriptsize±1.3  & 26.3\scriptsize±1.0 & 14.4\scriptsize±0.9  & 16.4\scriptsize±1.4  & 19.5\scriptsize±1.5  & 21.6\scriptsize±1.4  \\\hline
UER (ours)   & \underline{41.5\scriptsize±1.4}& \underline{49.2\scriptsize±1.7}                & \textbf{55.8\scriptsize±1.9}                 & \textbf{60.3\scriptsize±1.6} & \textbf{20.9\scriptsize±0.8} & \textbf{24.6\scriptsize±0.8}                &\textbf{27.0\scriptsize±0.5}                  & \textbf{29.6\scriptsize±1.1} &  \textbf{18.4\scriptsize±0.8}& \textbf{21.9\scriptsize±1.3}                   & \textbf{25.1\scriptsize±1.1}                  & \textbf{27.5\scriptsize±1.1} \\
$\hookrightarrow$UER-A  & 41.0\scriptsize±1.5& 47.6\scriptsize±1.4                & 51.9\scriptsize±1.9                 &55.5\scriptsize±1.8 & 18.6\scriptsize±0.6& 21.5\scriptsize±0.7                &24.5\scriptsize±0.6                  &  26.5\scriptsize±0.9 &  15.1\scriptsize±0.8& 17.8\scriptsize±1.4                   & 20.8\scriptsize±1.3                  & 22.6\scriptsize±1.3 \\
\hline
\end{tabular}
\vspace{-2ex}
\end{table*}

\begin{algorithm}[t]
\caption{Unbias Experience Replay}
\label{alg:lsr}
\renewcommand{\algorithmicrequire}{\textbf{Input:}}
\renewcommand{\algorithmicensure}{\textbf{Output:}}
\begin{algorithmic}[1]
\REQUIRE Dataset $D$, Learning Rate $\lambda$, Trade-off Parameter $\alpha$
\ENSURE  Network Parameters $\bm{\theta}$
\STATE \textbf{Initialize}: Memory Buffer $\mathcal{M}\leftarrow\{\}$
\FOR{$t\in\{1,2,...,T\}$}
\STATE //$Training\ Phase$
    \FOR{$\mathcal{B}_t\in D$}
        \STATE$\mathcal{B}_\mathcal{M}\leftarrow MemoryRetrieval(\mathcal{M})$.
        \STATE$L_{c}\leftarrow E_{(\bm{x},y)\sim{\mathcal{B}_t}}[l(p^{cos}(\bm{x}),\bm{y})]$
        \STATE$L_{p}\leftarrow E_{(\bm{x},y)\sim{\mathcal{B}_\mathcal{M}}}[\alpha l(p^{dot}(\bm{x}),\bm{y})+(1-\alpha)l(p^{cos}(\bm{x}),\bm{y})]$
        \STATE$L\leftarrow L_{c}+L_{p}$   
        \STATE$\bm{\theta}\leftarrow\bm{\theta}+\lambda\nabla_{\bm{\theta}}L$.
        \STATE$\mathcal{M}\leftarrow MemoryUpdate(\mathcal{M},\mathcal{B}_t)$.
    \ENDFOR
\STATE //$Testing\ Phase$
\STATE $m\leftarrow number\ of\ testing\ samples$
    \FOR{$i\in\{1,2,...,m\}$}
    \STATE $\hat{y} \leftarrow \mathop{\arg\max}_{c}p^{dot}(\bm{x}_i)[c],c\in \mathcal{C}_{t}$
    \ENDFOR
\STATE \textbf{return} $\theta$
\ENDFOR
\end{algorithmic}
\end{algorithm}

\subsection{Online Continual Training}
In this part, it trains the model by calculating logits in different ways when learning current samples and replaying previous samples. Based on the initialized model, the framework not only takes the angle factor to save novel knowledge but also leverages the norm factor to enhance all of learned knowledge. It can address the bias issue of prediction and further alleviate the phenomenon of CF by optimizing the objective function as
\begin{equation}
\begin{split}
    \label{eq:hcer_v3}
    L=L_{c}+L_{p},
\end{split}
\end{equation}

\noindent which has two components to learn training samples.

\textbf{The learning component} $L_{c}$ is the general cross-entropy loss function based on cosine logits, which is only associated with current samples. The target of this component is to mine novel knowledge of new classes, saving it into the angle factor. And the loss value is calculated as follows

\begin{equation}
\begin{split}
    \label{eq:first_term}
    L_{c}=E_{(\bm{x},y)\sim{\mathcal{B}_t}}[l(p^{cos}(\bm{x}),\bm{y})].
\end{split}
\end{equation}

\textbf{The replaying component} $L_{p}$ combines different cross-entropy loss functions. It is only related to the previous samples in the memory buffer, and it calculates categorical probability in both angle factor and norm factor. The loss with dot-product logits leverages norm factor to enhance historical knowledge and deal with the bias issue. In the meantime, the loss with cosine logits can control the strength of leveraging and prevent the model from losing its generalization ability. In detail, the framework takes $p^{dot}(\bm{x})$ as the categorical probability of previous samples, which can bring more postitive changes to parameters of previous classes. Besides, it takes $p^{cos}(\bm{x})$ of previous samples to balance the strength of the change. Finally, it can get the objective function as

\begin{equation}\small
    \label{eq:second_term}
    \resizebox{0.91\linewidth}{!}{$
            L_{p}=E_{(\bm{x},y)\sim{\mathcal{B}_\mathcal{M}}}[\alpha l(p^{dot}(\bm{x}),\bm{y})+(1-\alpha)l(p^{cos}(\bm{x}),\bm{y})],
        $}
\end{equation}%

\noindent where $\alpha$ is the hyper-parameter of leveraged scale.

\subsection{Online Continual Testing}
The \textbf{testing component} is similar to the replaying process of previous samples. Each testing sample $\bm{x}$ propagates forward through the network to obtain its class probability distribution $p^{dot}(\bm{x})$. Then we can classify it based on the following formula
\begin{equation}
 	\begin{split}
          \label{eq:third_term}
	 	\hat{y} = \mathop{\arg\max}_{c}p^{dot}(\bm{x})[c],c\in \mathcal{C}_{t}.
 	\end{split}
\end{equation}

The whole training and inference procedures are summarized in Algorithm~\ref{alg:lsr}. To begin with, the part of memory buffer operation set a fixed-size memory buffer to save (line 10) and replay previous samples (line 5). Then, the part of online continual training (line 6-9) overcomes the forgetting problem by learning current samples and previous samples by different objective functions, respectively. Finally, the part of online continual testing (line 13-16) predicts the categorical probability of unknown sample by dot-product logits, which is helpful to correctly classify the testing samples.

\section{Experiments}
\label{sec:experiments}

\subsection{Experiment Setup}
\subsubsection{Datasets}

To evaluate the effectiveness of our replay-based framework UER, we conduct extensive experiments on three publicly available datasets (CIFAR10~\cite{krizhevsky2009learning}, CIFAR100~\cite{krizhevsky2009learning} and MiniImageNet~\cite{vinyals2016matching}). \textbf{Split CIFAR10} consists of 5 learning stages and each stage contains 2 classes. \textbf{Split CIFAR100} and \textbf{Split MiniImageNet} are made up of 10 learning stages and each stage has 10 classes.

\begin{figure*}[t]
\centering
\subfigure[Split CIFAR10]{
    \begin{minipage}[t]{0.31\linewidth}
        \centering
        \includegraphics[scale=0.2]{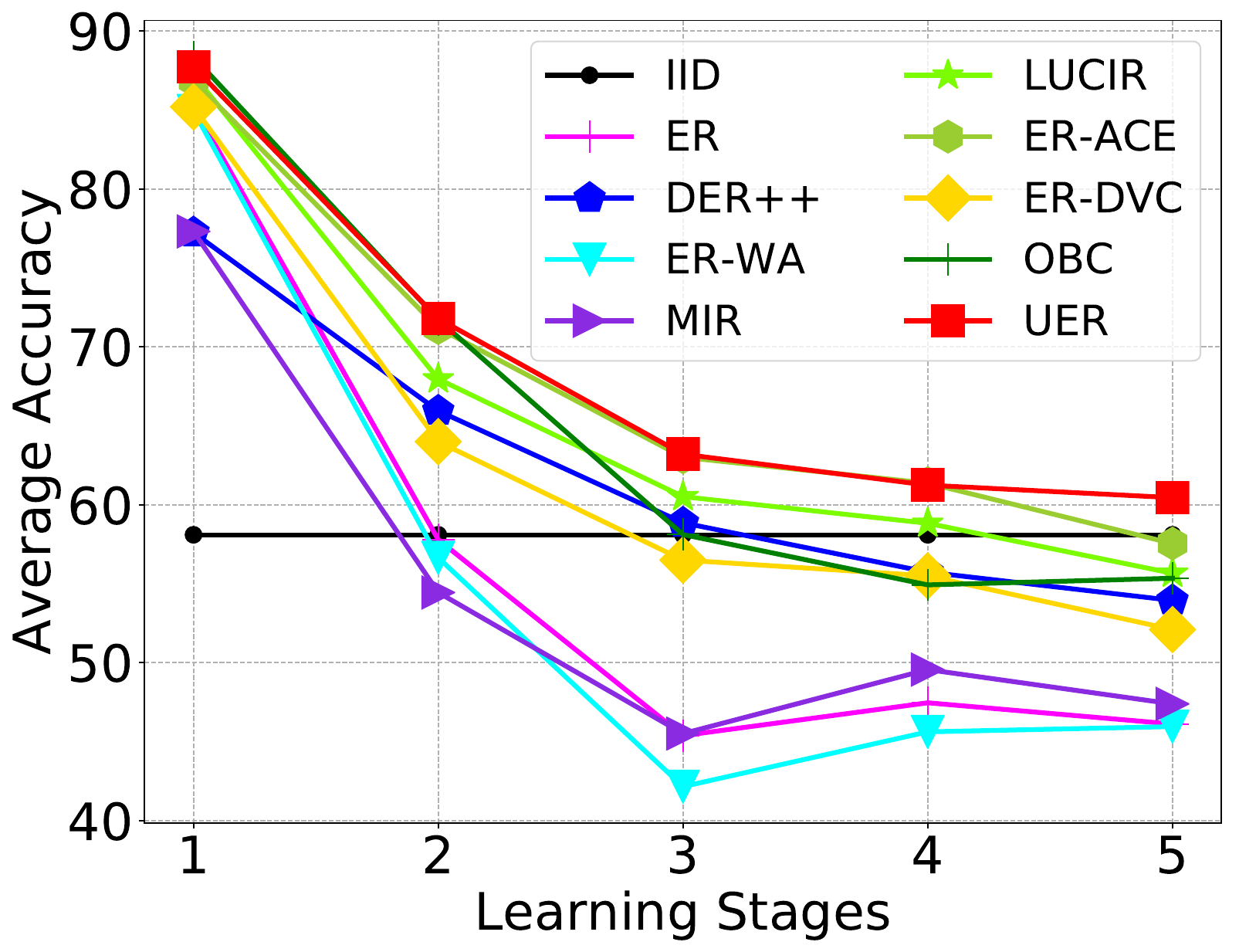}
    \end{minipage}
}
\subfigure[Split CIFAR100]{
    \begin{minipage}[t]{0.31\linewidth}
        \centering
        \includegraphics[scale=0.2]{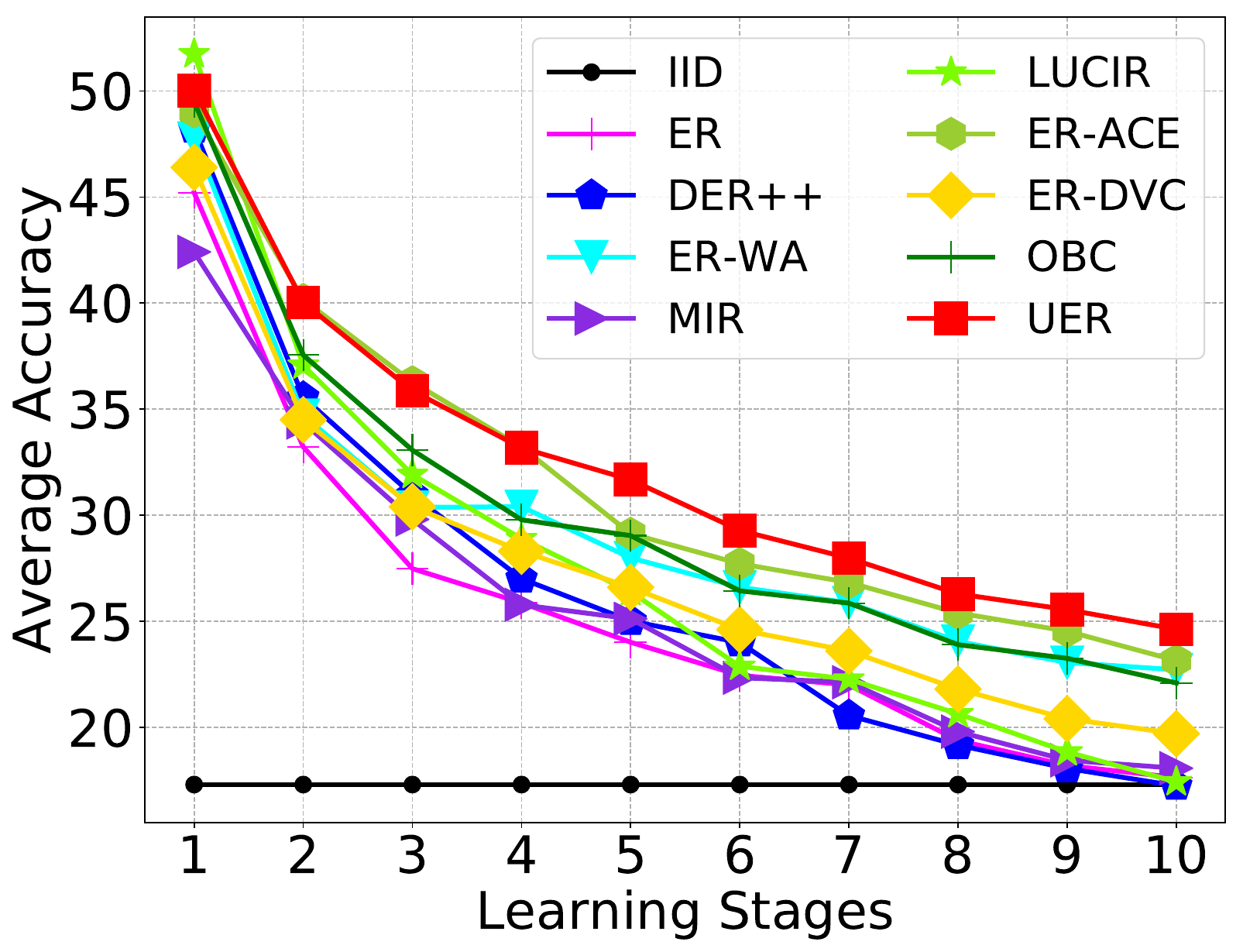}
    \end{minipage}
}
\subfigure[Split MiniImageNet]{
    \begin{minipage}[t]{0.31\linewidth}
        \centering
        \includegraphics[scale=0.2]{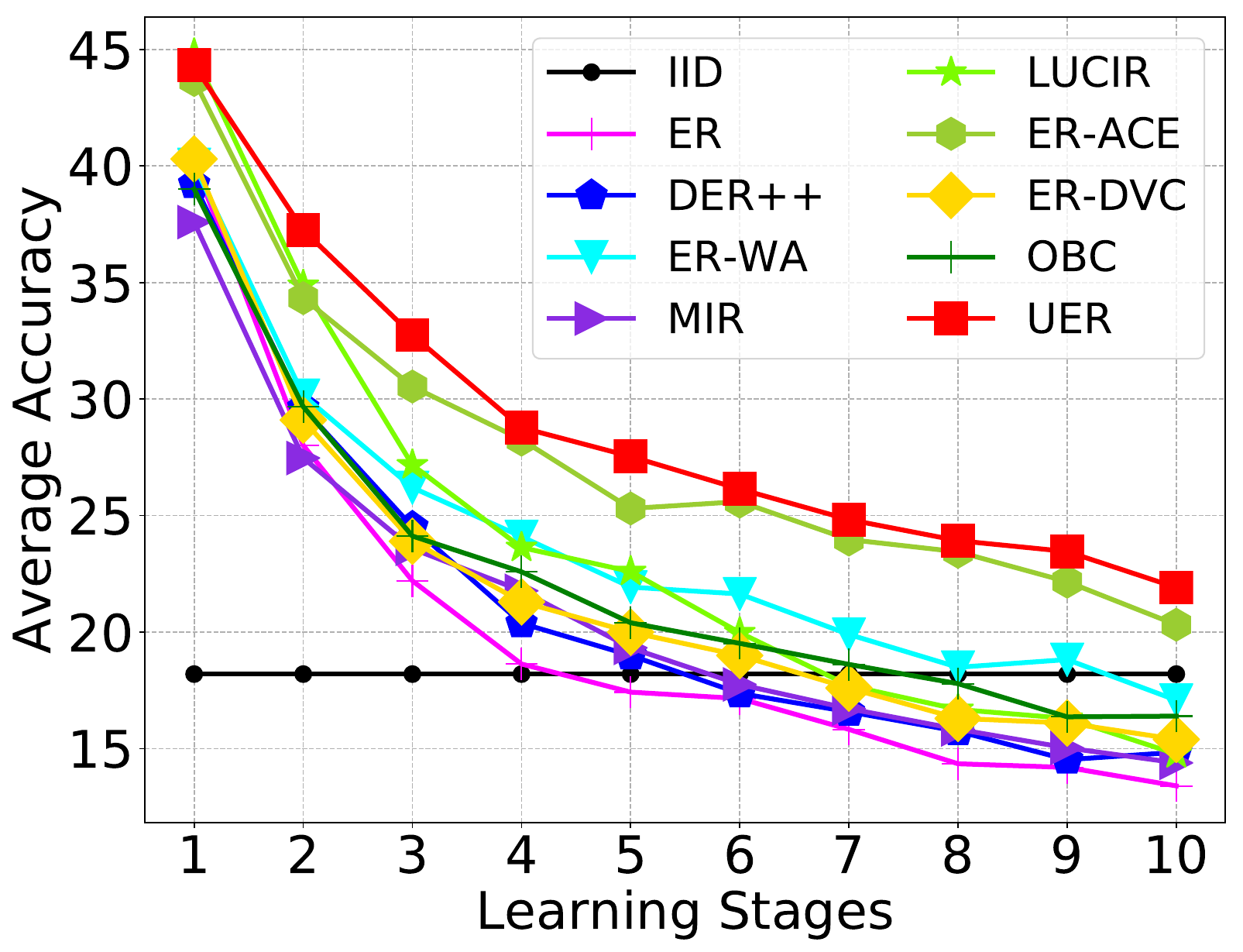}
    \end{minipage}
}
\centering
\vspace{-4ex}
\caption{Average accuracy rate on observed learning stages on three datasets when the memory buffer size is 1000. }
\label{learningstage1000}
\vspace{-1ex}
\end{figure*}


\subsubsection{Evaluation Metrics}
We define $a_{i,j}(j<=i)$ as the accuracy evaluated on the held-out test samples of the $j$th stage after the network has learned the first $i$ stages. Similar with~\cite{shim2021online}, we can acquire average accuracy rate as
\begin{equation}
\label{eq:first}
 	\begin{split}
	 	A_i = \frac{1}{i}\sum_{j=1}^{i}a_{i,j}.
 	\end{split}
\end{equation}

\subsubsection{Implementation Details}

The basic setting of backbone model is the same as the existing work~\cite{caccia2021new}. In detail, we take the Reduced ResNet18 (the number of filters is 20) as the feature extractor on all datasets. All of the networks are randomly initialized rather than pre-trained. During the training phase, the network is trained with SGD optimizer and learning rate is set as 0.1. Meanwhile, we select $\alpha$ on a validation set that obtained by sampling 10\% of the training set. As for the testing phase, we select 256 as the batch size. 

For all datasets, the class indexes are shuffled before division. The model receives 10 current samples from data stream and 10 previous samples from the memory buffer at a time irrespective of the size of the memory. Moreover, we employ a combination of various augmentation techniques to get the augmented images.

\begin{figure*}[t]
\centering
    \begin{minipage}[t]{0.49\linewidth}
        \centering
        \includegraphics[scale=0.24]{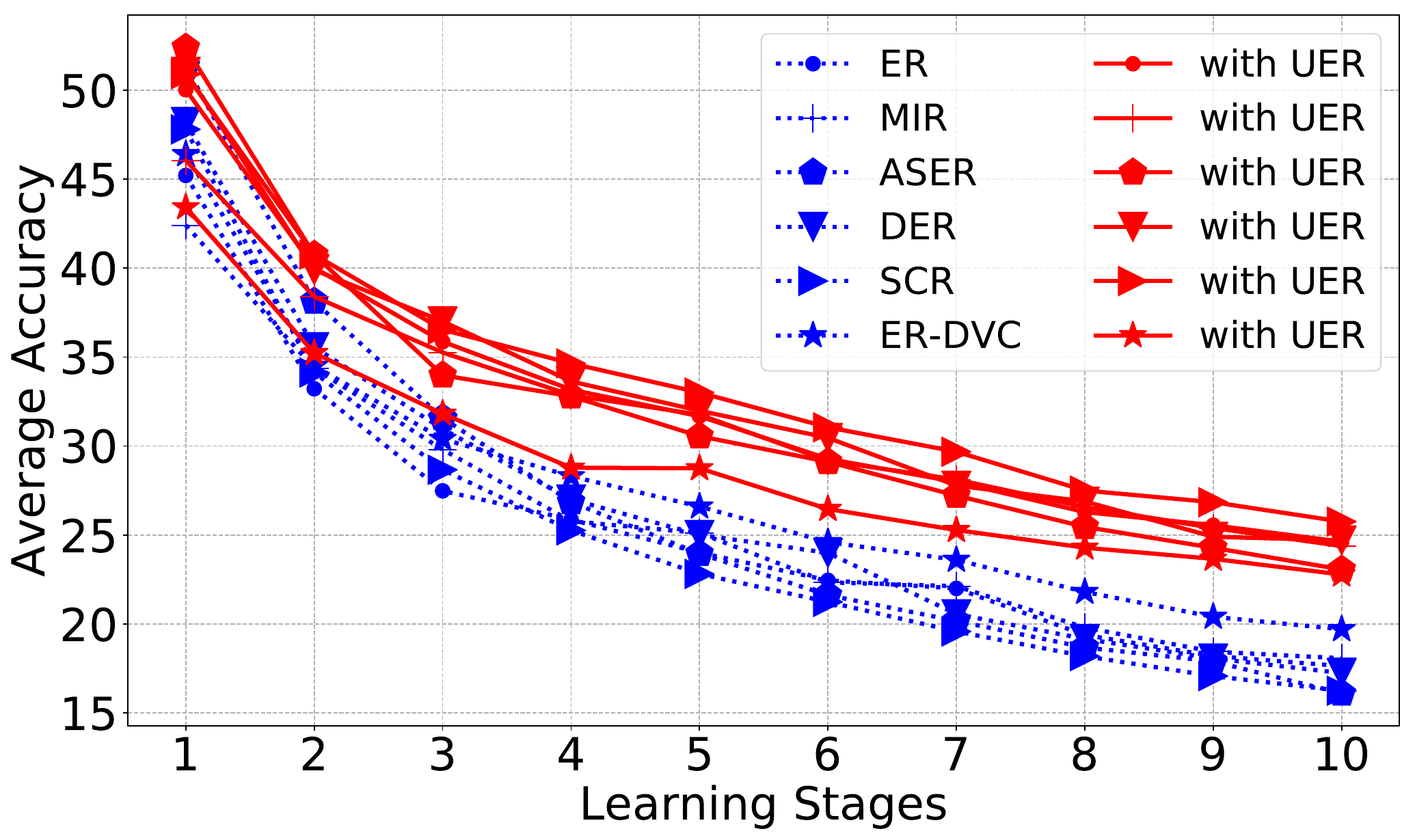}
        {Split CIFAR100}
    \end{minipage}
    \begin{minipage}[t]{0.49\linewidth}
        \centering
        \includegraphics[scale=0.24]{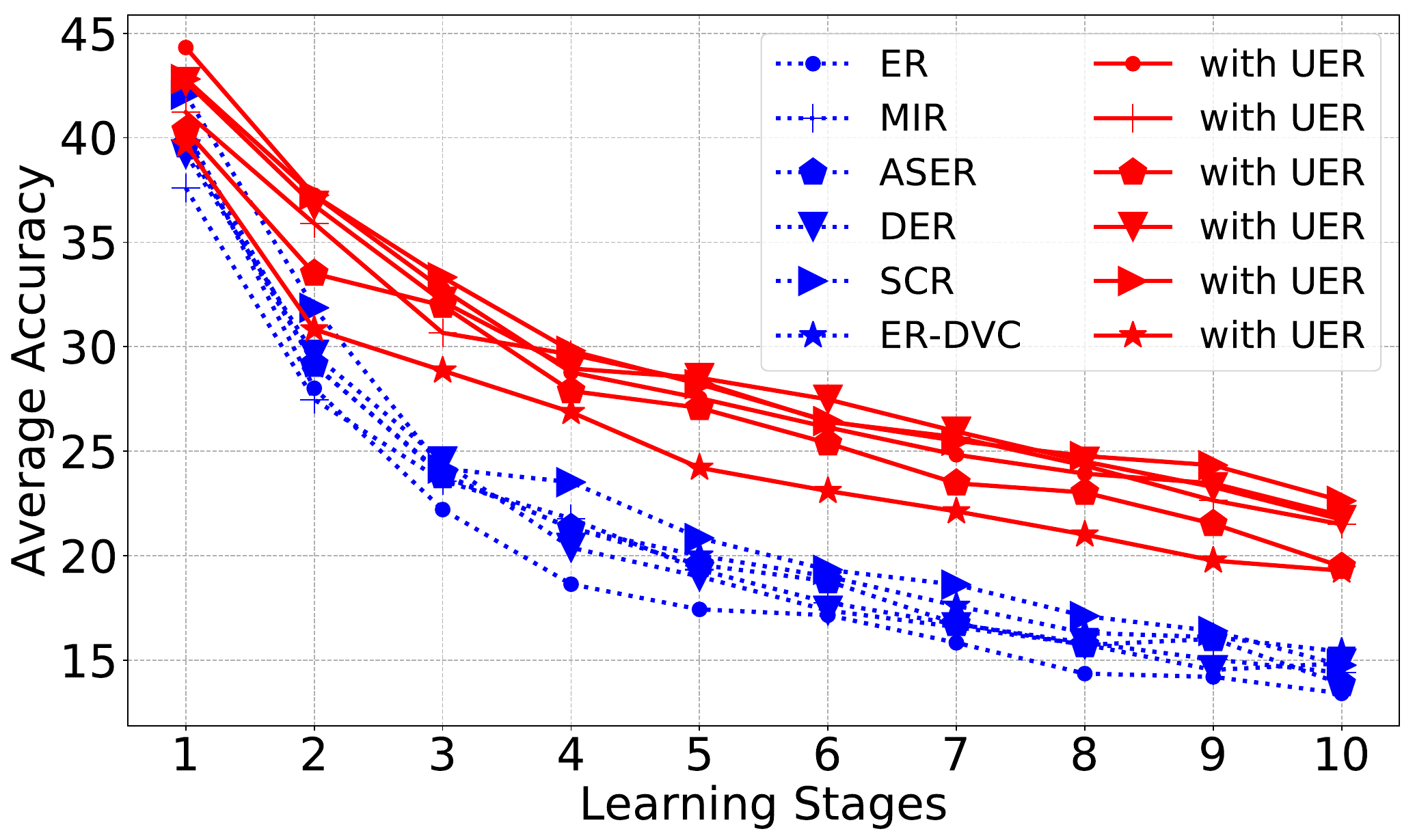}
        {Split MiniImageNet}
    \end{minipage}
\centering
\vspace{-2ex}
\caption{Improvement of existing methods with UER on Split CIFAR100 and Split MiniImageNet while the buffer size is 1000. }
\label{improvement}
\vspace{-2ex}
\end{figure*}

\subsection{Overall Performance}
In this section, we conduct experiments for UER and various state-of-the-art baselines on three datasets. We not only directly compare the performance of UER with these methods, but also explore the improvement of UER for them.

\textbf{Comparison with existing baselines.} Table~\ref{tableaccuracy} demonstrates the final average accuracy for three datasets. All reported scores are the average of 10 runs with a 95\% confidence interval. Among all rehearsal-based methods, the model update strategies are the most effective, and ER-ACE method achieves the best results.

Nevertheless, our proposed UER still outperforms them, demonstrating its effectiveness. Specifically, UER achieves the best performance under 10 of the 12 experimental scenarios, where three datasets contain four memory buffer of different sizes. It has the most outstanding performance on Split CIFAR100 and Split MiniImageNet. Meanwhile, the growth of buffer size further improves the performance of the UER. For example, UER outperforms the strongest baseline ER-ACE with a gap of 0.3\%, 1.6\%, 0.3\%, 1.3\% on Split MiniImagenet when the size of memory buffer is 500, 1000, 2000 and 5000, respectively. Moreover, UER defeats the ER-ACE with an improvement of 1.2\%, 1.5\%, 2.2\% and 2.6\% on Split CIFAR100 with 500, 1000, 2000 and 5000 size of memory buffer, respectively. We note that the performance of UER is not optimal on Split CIFAR10 when the buffer size is smaller. It is because there are fewer classes in this dataset, and the bias issue is not very serious. Meanwhile, the learning of norm factor is insufficient when the number of samples is small.

We report the accuracy performance at each stage for some of effective approaches on all datasets. As described in Figure~\ref{learningstage1000}, UER not only achieves significant results in the accuracy of final stage, but also consistently outperforms other baselines throughout the entire learning process. Its effectiveness becomes more and more visible as the stages increase, exhibiting its capability to overcome CF. For instance, although the advantage of UER is not obvious at the second stage and the third stage, its performance of anti-forgetting is the best in the remaining stages on Split CIFAR100.

\textbf{Improvement for existing baselines.} We apply UER to existing baselines, and the results demonstrate that UER greatly improves their performance. Figure~\ref{improvement} states the performance of six ER-based baselines with and without the combination of our method on Split CIFAR100 and Split MiniImagenet. The blue lines denote the original accuracy of existing studies, while the red ones are the improved accuracy of them after combining with UER. By leveraging the norm factor to deal with the bias problem, the performance of these baselines in anti-forgetting is enhanced. For instance, the SCR that performs poorly for mini learning batches, achieves the best performance in the whole learning process with the help of our method. In summary, our method not only shows significant advantages over other baseline approaches, but can also be integrated with existing methods to greatly improve their performance. 

\begin{figure}[t]
\centering
\includegraphics[scale=0.22]{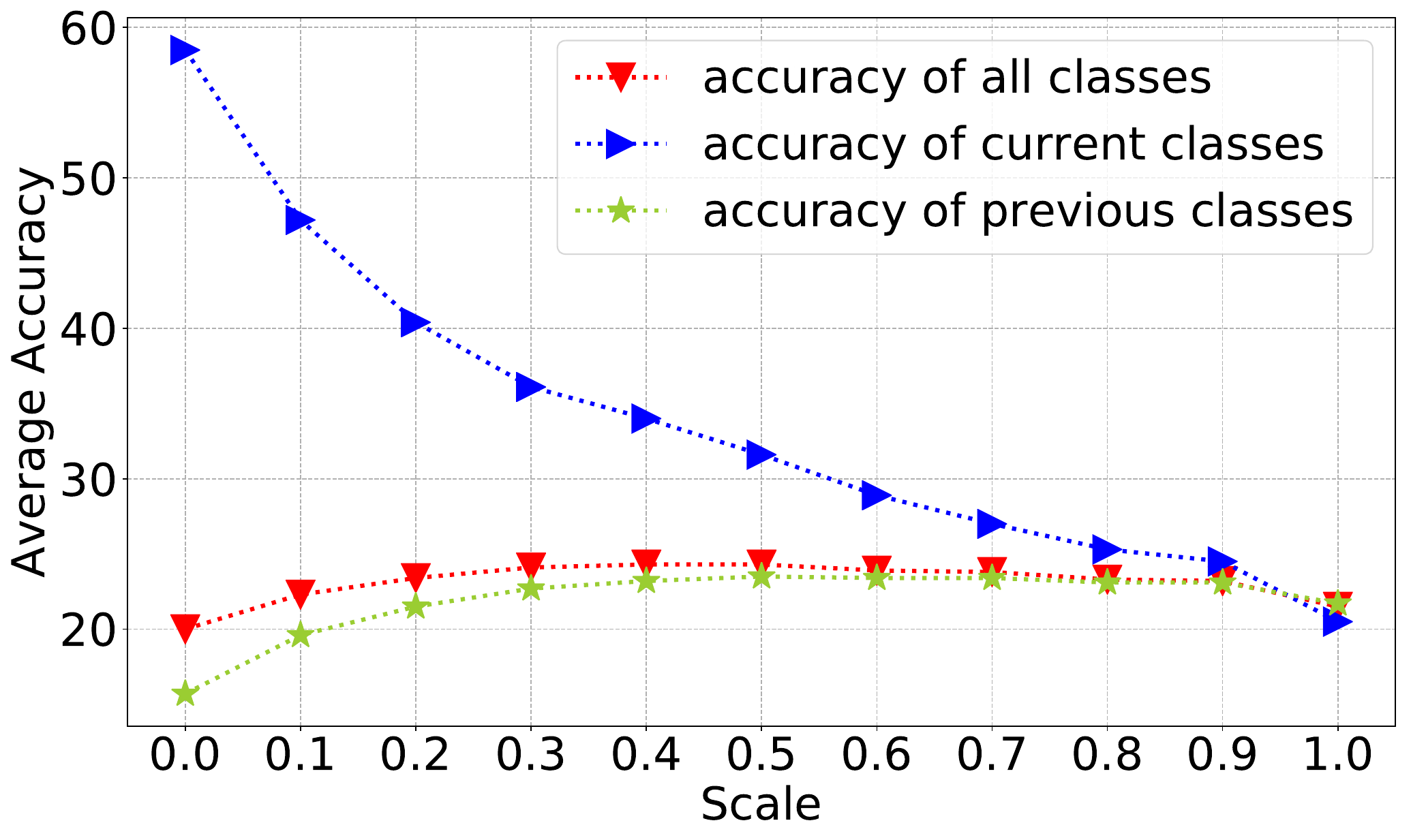}
\vspace{-2ex}
\caption{The performance of UER on Split CIFAR100 (buffer size=1000) with different values of $\alpha$.}
\label{fig:scale}%
\vspace{-4ex}
\end{figure}

\begin{figure*}[t]
\centering
\subfigure[Average of Previous Classes and Current Classes]{
    \begin{minipage}[t]{0.31\linewidth}
        \centering
        \includegraphics[scale=0.2]{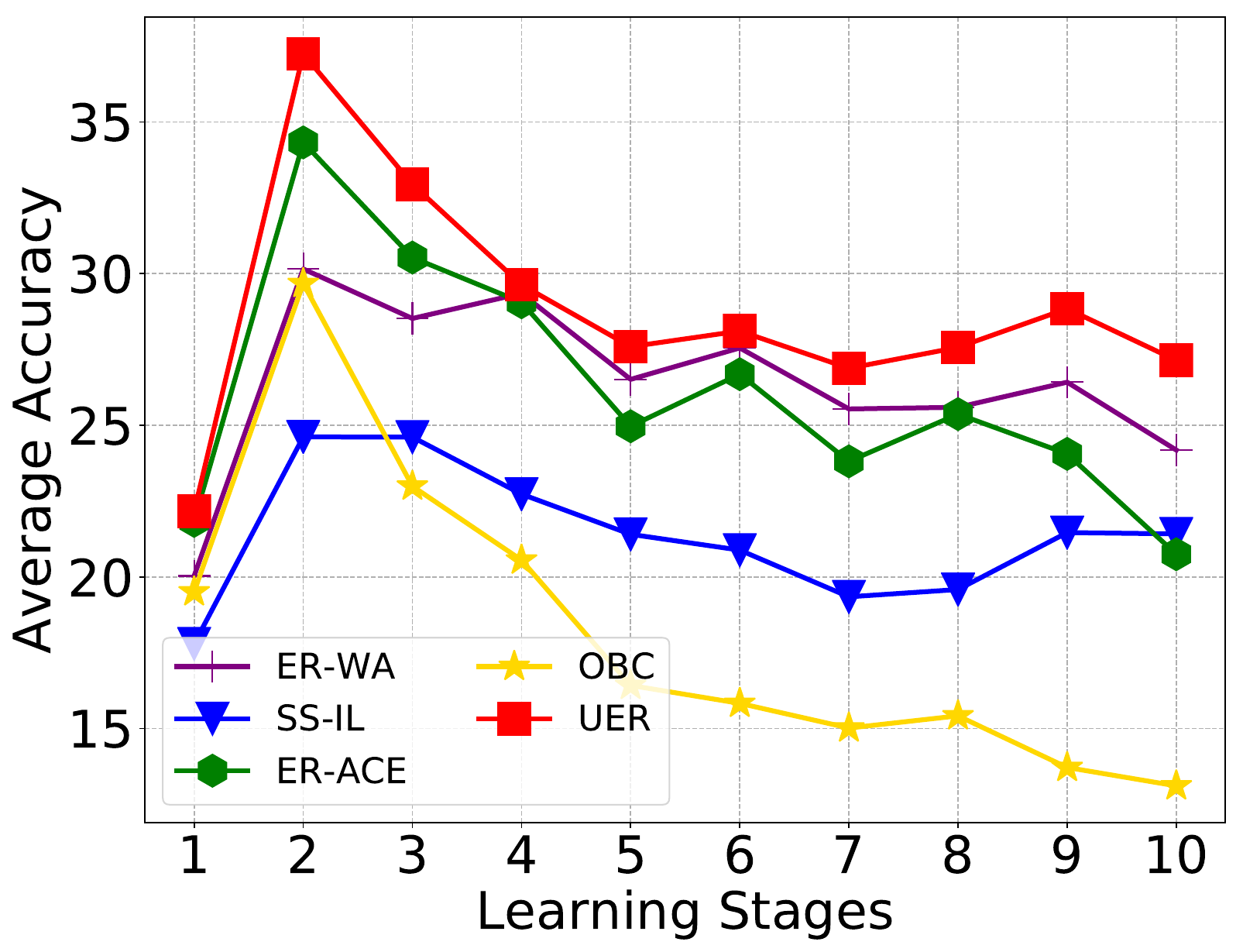}
    \end{minipage}
}
\subfigure[Previous Classes]{
    \begin{minipage}[t]{0.31\linewidth}
        \centering
        \includegraphics[scale=0.2]{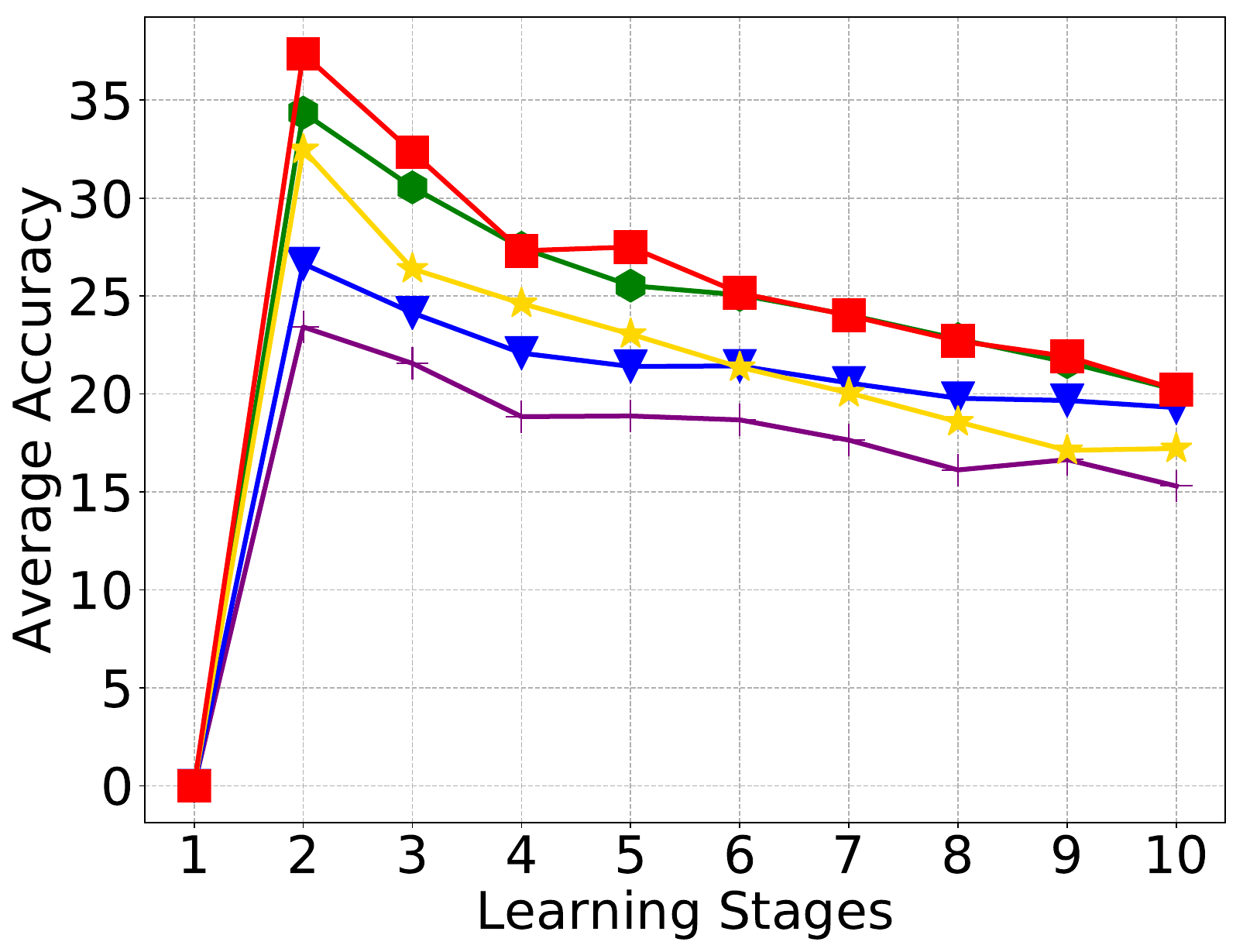}
    \end{minipage}
}
\subfigure[Current Classes]{
    \begin{minipage}[t]{0.31\linewidth}
        \centering
        \includegraphics[scale=0.2]{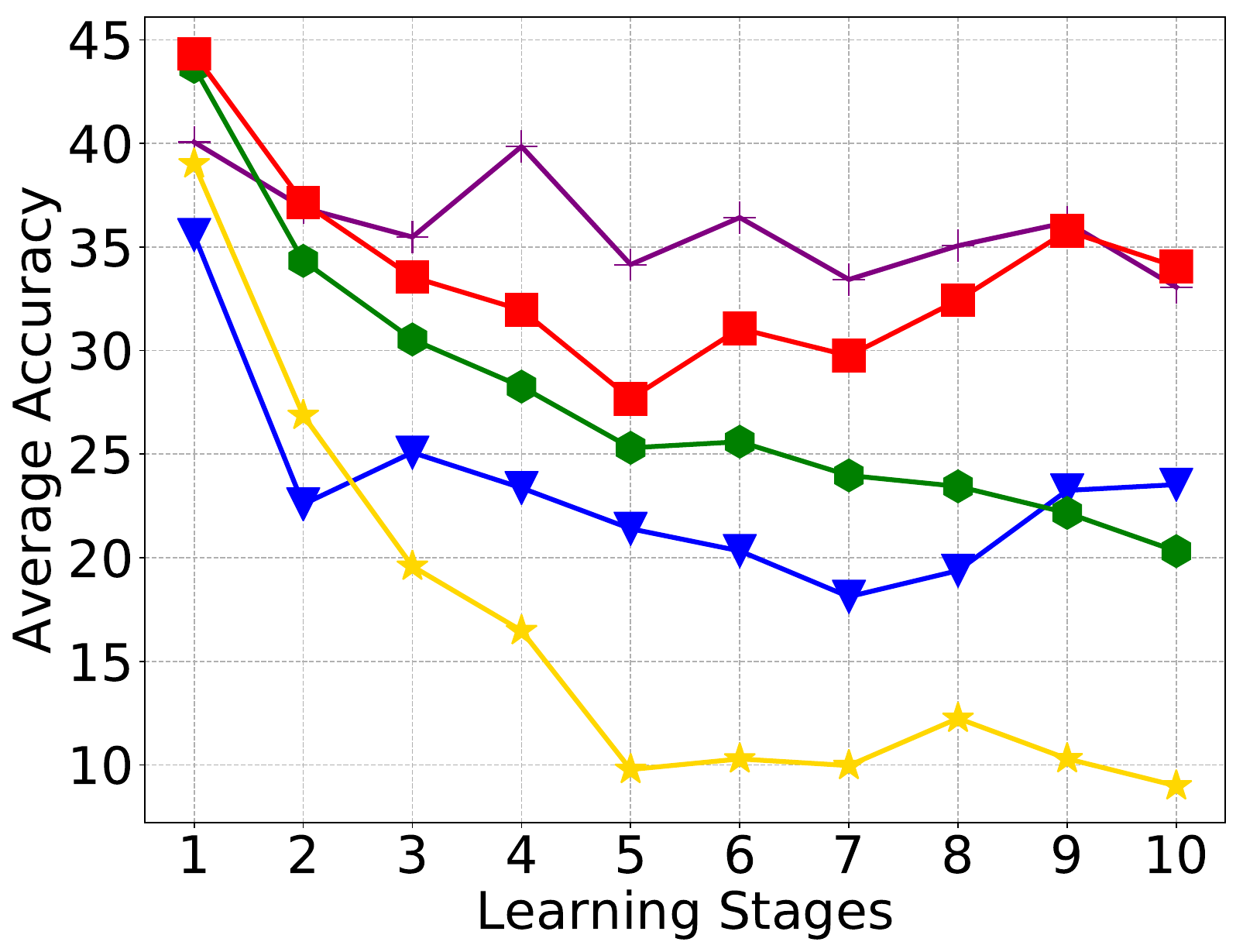}
    \end{minipage}
}
\centering
\vspace{-4ex}
\caption{Average accuracy rate of previous classes, current classes and their average on observed learning stages of Split MiniImageNet when the memory buffer size is 1000. }
\label{fig:balance}
\vspace{-2ex}
\end{figure*}

\subsection{Analysis Study}

\subsubsection{Real-time Ability}
UER is a simple and effective method for real-time prediction of OCL.
Similar with~\cite{ghunaim2023real}, we use $C_\mathcal{D}$ to compare the computation of different methods. The $C_\mathcal{D}$ is denoted as a
relative complexity between the stream and an underlying continual learning method. For example, when current samples come, ER can immediately learn them and then predict unknown samples, resulting in a relative complexity of 1. Since UER and ER-ACE only
modify the loss objective of ER, their computational complexities are equivalent to 1. However, OBC has to replay another batch of previous samples based on ER, making it require 1.5× FLOPs needed by ER. Thus, its computational complexity is 1.5, which limits the ability of real-time prediction. The comparison of running time between various methods also confirms this result.

\begin{table}[t]
\renewcommand\tabcolsep{2pt}
\centering
\caption{The computational complexity (lower is better) of the model on Split CIFAR100 (buffer size=1000).}
\vspace{-2ex}
\label{table:ablation}
\begin{tabular}{l|cccc}
\hline
Metric       & ER      & ER-ACE & OBC & UER\\ \hline
$C_\mathcal{D}$           & 1      & 1 & 1.5 & 1\\
Time/s       & 1740      & 1698 & 2542 & 1763\\
\hline
\end{tabular}
\vspace{-2ex}
\end{table}

\subsubsection{Flexibility}
Meanwhile, UER is flexible since it is not only suitable for the linear layer of predictor, but also beneficial for the one in feature extractor. Specifically, we add a linear layer to the feature extractor. The coupling module is used in the feature extractor, the predictor and all linear layers, respectively. Table~\ref{table:ablation} reports the final average accuracy rate of all classes, current classes and previous classes for all methods. It is not difficult to find that this module can work on each linear layer alone, and the overall performance is better and more stable on both linear layers. Hence, compared with other existing methods, UER is not constrained by the classes boundary, and has a wider range of applications.

\begin{table}[t]
\renewcommand\tabcolsep{2pt}
\centering
\caption{The Final Average Accuracy Rate (higher is better) of the model on Split CIFAR100 (buffer size=1000) when it adds a linear layer to the feature extractor.}
\vspace{-2ex}
\label{table:ablation}
\begin{tabular}{l|ccccc}
\hline
Method       & All Classes     & Current & Previous\\ \hline
ER           & 15.6\scriptsize±2.6 & 50.5      & 11.7 \\
ER-ACE       & 18.9\scriptsize±0.9 & 17.3      & 19.1 \\
OBC       & 18.3\scriptsize±0.7 & 17.5      & 18.4  \\
UER (feature extractor) & 18.6\scriptsize±1.0 & 30.6      & 17.2  \\
UER (predictor)  & 19.6\scriptsize±1.5 & 34.7      & 18.0  \\
UER (both)    & \textbf{20.1\scriptsize±1.2} & 31.8      & 18.8 \\ \hline
\end{tabular}
\vspace{-4ex}
\end{table}

\subsubsection{Adjustability}
By using dot-product logits, UER leverages the norm factor to balance the performance of current and previous classes. And the effect is mainly affected by the leveraged scale hyper-parameter $\alpha$. As shown in Table~\ref{tableaccuracy}, the ablation version of UER is denoted as UER-A. Due to the lack of scale adjustment, UER-A performs worse than UER.

To explore the impact of $\alpha$, we conduct experiments on Split CIFAR100 for different $\alpha$. In Figure~\ref{fig:scale}, we report the accuracy of all classes for UER, which can be divided into the accuracy of previous classes and the accuracy of current classes. As $\alpha$ increases, the proportion of dot-product logits raises. Its performance (red line) will be improved at the beginning, and it will not enhance or even start to decline when $\alpha$ becomes large. By comparing the performance of current and previous classes, we find that the accuracy of current classes (blue line) will decline with the increase of $\alpha$. While the accuracy rate of previous classes (green line) will rise at the beginning and then decline later. Emprically, UER can play its greatest role for the model when the value of $\alpha$ is around 0.5.

Hence, UER is more adjustable than other methods. In Figure~\ref{fig:balance}, we demonstrate the performance of several bias addressing methods on previous classes, current classes and their average. With the help of the scale, UER not only ensures the generalization ability to learn current classes, but also improves the memory ability to capture previous classes.   
\section{Conclusion}
\label{sec:conclusion}
In this paper, we develop a heuristic OCL method called UER to alleviate the phenomenon of CF by tackling the bias problem of logits. By examining the change of predictor, we find that the important factors of dot-product logits include a norm factor and an angle factor. The former is beneficial to previous classes while the latter is fond of current classes. As a result, the cosine logits depends on the angle factor is suitable to adapt novel knowledge. And the dot-product logits made up of norm and angle factors is good at storing historical knowledge. Hence, the proposed UER leverages the norm factor to balance the categorical prediction of the predictor. It learns current samples only by the angle factor and further replays previous samples by both the norm and angle factors. Extensive experiments on three datasets demonstrate the superiority of UER over various state-of-the-art baselines.

\begin{acks}
This work was supported in part by National Nature Science Foundation of China (No. 61972111, No. 62202124 and No. 62272130), Nature Science Program of Shenzhen (No. JCYJ20210324120208022 and No. JCYJ20200109113014456) and Shenzhen Science and Technology Program (No. KCXFZ20211020163403005). 
\end{acks}


\bibliographystyle{ACM-Reference-Format}
\balance



\end{document}